\documentclass[
twocolumn,
]{ceurart}

\usepackage[utf8]{inputenc} 
\usepackage[T1]{fontenc}    
\usepackage{hyperref}       
\usepackage{url}            
\usepackage{booktabs}       
\usepackage{amsfonts}       
\usepackage{nicefrac}       
\usepackage{microtype}      
\usepackage{xcolor}         
\usepackage{graphicx}       
\usepackage{float}          
\usepackage{stfloats}          
\usepackage{graphicx} 

\usepackage{amsmath,amssymb,amsthm}
\usepackage{centernot}
\usepackage{parskip}
\usepackage{bm}
\usepackage{natbib}
\usepackage{listings}
\usepackage{multirow}

\theoremstyle{definition}

\begin{document}

\copyrightyear{2024}
\copyrightclause{Copyright for this paper by its authors.
  Use permitted under Creative Commons License Attribution 4.0
  International (CC BY 4.0).}

\conference{KiL'24: Workshop on Knowledge-infused Learning co-located with 30th ACM KDD Conference,
August 26, 2024, Barcelona, Spain}

\title{GraphEval: A Knowledge-Graph Based LLM Hallucination Evaluation Framework}


\author[1]{Hannah Sansford}[%
email=hannah.sansford@bristol.ac.uk,
]
\cormark[1]
\address[1]{University of Bristol, UK}

\address[2]{Amazon Science}

\author[2]{Nicholas Richardson}[%
email=nchls@amazon.es,
]

\author[2]{Hermina {Petric Maretic}}[%
email=maretich@amazon.co.uk,
]

\author[2]{Juba {Nait Saada}}[%
email=jubans@amazon.co.uk,
]

\cortext[1]{Work done during an internship with Amazon.}

\begin{abstract}
    Methods to evaluate Large Language Model (LLM) responses and detect inconsistencies, also known as hallucinations, with respect to the provided knowledge, are becoming increasingly important for LLM applications. Current metrics fall short in their ability to provide explainable decisions, systematically check all pieces of information in the response, and are often too computationally expensive to be used in practice. We present GraphEval: a hallucination evaluation framework based on representing information in Knowledge Graph (KG) structures. Our method identifies the specific triples in the KG that are prone to hallucinations and hence provides more insight into where in the response a hallucination has occurred, if at all, than previous methods.
    Furthermore, using our approach in conjunction with state-of-the-art natural language inference (NLI) models leads to an improvement in balanced accuracy on various hallucination benchmarks, compared to using the raw NLI models. Lastly, we explore the use of GraphEval for hallucination correction by leveraging the structure of the KG, a method we name GraphCorrect, and demonstrate that the majority of hallucinations can indeed be rectified.
\end{abstract}

\begin{keywords}
  Large Language Models \sep
  Knowledge Graphs \sep
  Hallucination Detection \sep
  Hallucination Correction
\end{keywords}

\maketitle

\section{Introduction}

As the size and power of LLMs have drastically increased over recent years, so has the number of potential applications. Arguably, one of the biggest blockers to implementing these models in practice is their tendency to hallucinate - returning seemingly plausible, but untrue, responses. Here, we focus on the problem of detecting hallucinations with respect to the provided context that the LLM should use as its source of knowledge; detecting hallucinations that have deviated from the LLM's original training data is out of the scope of this work.
In applications where certainty in a response is critical, such as medical diagnosis, the existence of hallucinations that arise from a given context is especially limiting. Therefore, it is of utmost importance to develop successful methods to detect these hallucinations and, when it is of interest to address or correct them, provide clarity on which aspect of the response is likely a hallucination. The importance of this issue is reflected in the amount of research being published on the topic - see \citet{ji2023survey} for a recent survey of this area.

Performing evaluation on natural language is a challenging task that researchers have been interested in long before hallucinations were at the forefront of the problem. Methods have evolved a great deal from traditional N-gram based metrics, such as BLEU \citep{papineni-etal-2002-bleu} and ROUGE \citep{lin-2004-rouge}, to much more intricate LLM-based evaluation metrics with user-defined evaluation criteria, such as G-Eval \citep{liu-etal-2023-g}. More recently, techniques to mitigate the prevalence of hallucinations in generated outputs leveraging Retrieval Augmented Generation (RAG) \citep{lewis2020retrieval} and reasoning on knowledge graphs (KGs) \citep{luo2023reasoning, yang2024give} have been proposed. The former suggested the concatenation of relevant contextual data into the prompt to ground the LLM response, while the latter enforced a more robust reasoning process through providing grounding information in KG structures \citep{agrawal2024knowledge}. As successful as these approaches have been, they do not fully circumvent the need to evaluate LLM outputs.

Inspired by current research harnessing KGs to provide grounded LLM responses, we propose GraphEval - a hallucination detection framework based on the representation of information in KG structures. To the best of our knowledge, we are the first to apply KGs to an LLM-based hallucination evaluation framework, and in doing so we provide a higher level of insight into where in the output a hallucination has occurred than any previous metrics. Additionally, we demonstrate how using our method in conjunction with current state-of-the-art hallucination detection methods improves their classification accuracy on various benchmarks. Finally, we consider the problem of hallucination correction and we introduce GraphCorrect, showcasing how GraphEval can effectively be extended to rectify a significant proportion of hallucinations present in LLM outputs.





\section{Problem statement}

In this work we focus on the \textit{closed-domain} hallucination detection problem: the situation where we have a textual output from an LLM which is generated using some grounding context included in the prompt. In this case, the goal is for the LLM to use the provided context as its only source of knowledge. The \textit{open-domain} problem, which is with respect to all factual knowledge in the world, is not explored here but is briefly discussed in Section \ref{future_work}.

We consider hallucination detection to be a binary classification problem, with 0 corresponding to the LLM output being factually consistent given the provided context, and 1 corresponding to the output containing at least one inconsistency. We can assess hallucination evaluation methods using a benchmarking dataset containing ground-truth labels (usually human-annotated) to determine whether a given context-output pair contains factual inconsistencies. Throughout the paper we use the terms factual, consistent, grounded and faithful interchangeably to mean \textit{containing no hallucinations with respect to the context}.

Finally, we explore the problem of hallucination correction, wherein we do not use any directly labeled dataset. Instead, we utilize hallucination detection frameworks to first identify hallucinations to correct, and subsequently repurposing them to evaluate the corrected outputs. It is important to note that our exploration of hallucination correction only serves as an extension to our evaluation framework and is not the primary focus of this study.

\begin{figure*}
    \centering
    \includegraphics[width=1\linewidth]{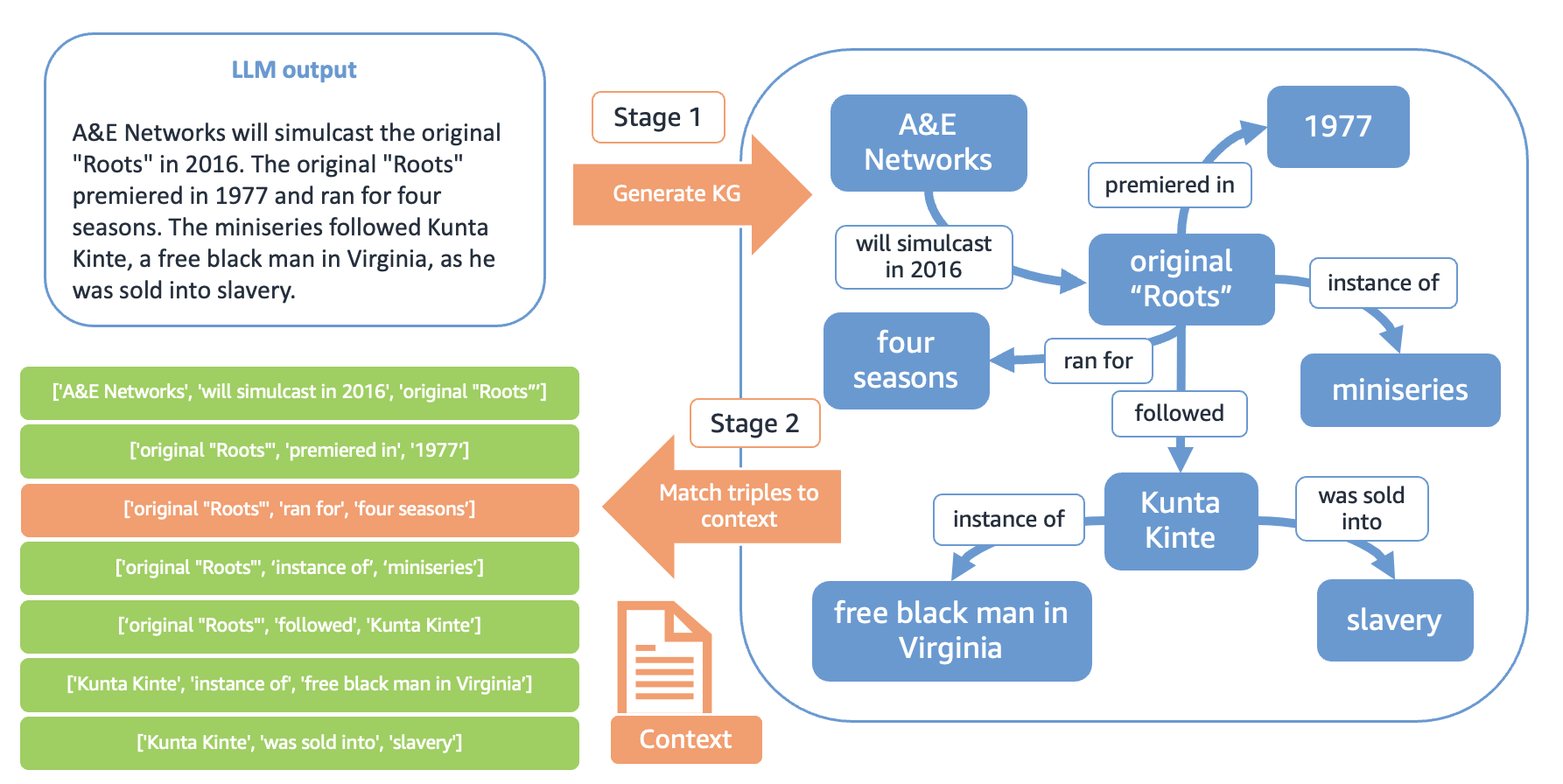}
    \caption{A visualisation of the GraphEval approach. First, the LLM output is fed into the KG construction prompt to produce the KG depicted on the right. Next, each individual triple in the KG is fed into an out-of-the-box hallucination detection method, such as an NLI model, and compared to the provided context for inconsistencies. Finally, any triples that are flagged as inconsistent are returned to the user, along with the overall hallucination decision.}
    \label{fig:grapheval}
\end{figure*}

\section{Related work}

Historically, N-gram based metrics such as BLEU \citep{papineni-etal-2002-bleu} and ROUGE \citep{lin-2004-rouge} have been the most widely used metrics for natural language evaluation. However, these metrics have been shown to perform poorly at the task of factual inconsistency detection \citep{maynez-etal-2020-faithfulness, honovich-etal-2021-q2}. In more recent years, embedding-based metrics such as BERTScore \citep{Zhang*2020BERTScore:} have been favoured over N-gram based metrics. These methods measure the similarity between two pieces of text by comparing the contextualised embedding from a transformer model, such as BERT \citep{devlin-etal-2019-bert}. 

Both N-gram and embedding-based metrics base their scores on how similar the text to be evaluated is to some reference text. This similarity objective often fails to capture the intricacies of the hallucination detection problem. Therefore, researchers have begun to develop new methods that are more acutely tuned to detecting inconsistencies between an LLM output and its grounding context. \citet{maynez-etal-2020-faithfulness} identified the crossover between the textual entailment score in NLI tasks and consistency prediction. This was a breakthrough at the time, producing higher correlation with faithfulness than any previous metrics, and paved the way for further research that capitalised on NLI data and models \citep{honovich-etal-2022-true, laban-etal-2022-summac, gekhman2023trueteacher}.

Very recently, attention has turned to leveraging LLMs themselves to evaluate the consistency of LLM outputs. SelfCheckGPT \citep{manakul2023selfcheckgpt} and ChatProtect \citep{mundler2024selfcontradictory} approach the problem by considering the self-consistency within sampled outputs.
Since they require the generation of a large number of responses from the LLM, many consider these methods prohibitively computationally expensive. 

Other LLM-based hallucination evaluation methods, such as G-Eval \citep{liu-etal-2023-g} and GPTScore \citep{fu2023gptscore}, employ a different LLM for evaluation than the one used to generate the LLM response that needs to be evaluated. G-Eval allows user-defined evaluation criteria and uses automated chain-of-thought prompting and form-filling to assign scores. GPTScore treats the task as conditional generation, leveraging models like GPT-3 to assign higher probabilities to high-quality outputs by prepending evaluation instructions to the LLM prompt. Unlike NLI models trained on binary classification data, these methods produce scores that are harder to interpret as probabilities and often require additional steps for inconsistency classification.

Recent hallucination detection methods, such as FactScore \cite{min2023factscore} and SAFE \cite{wei2024long}, utilize large language models to break down the response into \textit{atomic} or \textit{individual} facts for evaluation. These approaches have enabled precise identification of where hallucinations occur within the LLM response. Each fact is automatically verified against a comprehensive knowledge source like Wikipedia or scientific literature in the case of FactScore, or through the use of a search engine in the case of SAFE.

FactGraph \citep{ribeiro2022factgraph} is the only factuality evaluation method we are aware of that utilises graph-like structures. The method is focused solely on the detection of inconsistencies in the summarization problem, decomposing both the summary and the supporting documents into what they call structured \textit{meaning representations} (MRs). These MRs describe the core semantic concepts and relations, which the authors claim to be more suitable for factuality evaluation than the raw text.

\section{GraphEval: Our evaluation method}
\label{our_method}


GraphEval is based around the idea of representing information in a structured manner through KGs, and aims to address the lack of explainability of previous hallucination detection approaches, i.e. which concrete pieces of information in particular are inconsistent. 


Formally, a KG is a collection of triples $\mathcal{KG} = \{ (e_1, r, e_2) \subseteq \mathcal{E} \times \mathcal{R} \times \mathcal{E}\}$, where $\mathcal{E}$ and $\mathcal{R}$ denote the set of entities and relationships, respectively. In the GraphEval setting, both entities and relationships are simply pieces of text. We do not make use of common extensions to this simple setting, such as entity and relationship types, or attached properties.

Our GraphEval metric consists of a two-stage procedure:
\begin{itemize}
    \item[] \textbf{Stage 1} - Construct a KG from the LLM output to be evaluated.
    \item[] \textbf{Stage 2} - Iterate through each of the triples in the KG, identifying whether they are factually consistent given the provided context.
\end{itemize}

The output is considered factually inconsistent if any of the triples in stage 2 are identified as not grounded in the context. The inconsistent triple(s) may also be returned to provide explainability by highlighting where in the output the hallucination(s) has occurred. We provide a visualisation of this process in Figure \ref{fig:grapheval} using a real example from one of the benchmarks described in Section \ref{benchmarks}.

Regarding stage 1, we provide a short review of LLM-based KG construction methods in Section \ref{kg_construction}, along with results from our implementation. For stage 2, we leverage existing techniques and employ an out-of-the-box NLI model for this task. A benefit of this approach is that it gives us the opportunity to make a direct comparison between the performance of the raw NLI model and the model supplemented with our KG approach. In essence, our method is a pre-processing step, the output of which can be fed into any hallucination detection method; we choose NLI models as they are computationally cheap compared to LLM-based models, yet still achieve state-of-the-art results. By feeding each triple into an NLI model, along with the grounding context, we obtain a probability of containing a hallucination for each triple. Finally, we classify the example as inconsistent if at least one triple produces a probability greater than 0.5. 


Similar approaches to ours have been proposed in recent literature. SummaC \citep{laban-etal-2022-summac} also uses NLI-based models to detect inconsistencies in LLM-generated summaries. However, it distinguishes itself by segmenting both the context and the summary into their respective sentences, and then by passing each context-summary pair into the NLI model. This approach presents challenges in maintaining entity references across sentences; for instance, "John Doe" may only be referred to as "he" in another sentence. Similarly, FactScore \cite{min2023factscore} faces the same limitation. Our method circumvents this issue by organising entity relationships with a KG. 


While FactGraph \citep{ribeiro2022factgraph} also makes use of graph structures in their consistency evaluation process, the method differs from GraphEval in a few major respects. Firstly, their approach can only be applied to the summarisation problem; whereas GraphEval can easily be applied to various domains such as Summarisation, Question Answering, Common Sense Reasoning and many others. Secondly, FactGraph does not employ LLMs anywhere in their framework, missing out on recent advances in the field. Finally, their approach aims to decompose \textit{both} the LLM output and the provided context into the underlying core semantic concepts and relations, before comparing each of the graph structures. GraphEval, on the other hand, only represents the LLM output as a KG and aims to preserve as much of the information contained in the raw text as possible.


To summarise the advantages of GraphEval over previous methods:
\begin{itemize}
    \item We present a systematic way of checking all pieces of information contained in the LLM output.
    \item Our method only requires one call to an LLM, in the KG construction phase, and does not require the (usually) large context documents to be input, as in all previous LLM-based metrics. This makes GraphEval less computationally expensive than other LLM-based methods.
    \item Our method returns the specific triples that are not grounded in the context, providing explainability for the decision and identifying which section of the output should not be trusted. We leverage this feature for hallucination correction and propose a new method called GraphCorrect, described in Section \ref{hallucinations_correction}.
\end{itemize}

\def\arraystretch{1.3}
\begin{table*}
    \centering
    \begin{tabular}{ccccc}
        \hline
        \textbf{Benchmark} & \textbf{No. of Examples} & \textbf{Label Ratio} & \textbf{Avg Output len.} & \textbf{Avg Context len.}\\
        \hline
         SummEval & 1,600 & 33.2\% & 63 & 359\\
         QAGS-C  & 235 & 48.1\%  & 49 & 383\\
         QAGS-X  & 239 & 48.5\% & 18 & 318\\
    \end{tabular}
    \vspace{3pt}
    \caption{Statistics relating to the evaluation benchmarks used. The label ratio is the ratio of factually consistent examples to inconsistent examples. The average output and context length are the average number of words in each.}
    \label{tab:benchmarks}
\end{table*}

\section{Construction of KGs using LLMs}
\label{kg_construction}

Constructing KGs from unstructured textual data involves identifying the set of entities within the text and the relationships between them, resulting in a structured representation of the information contained within the text. The process can be divided into three main stages:
\begin{enumerate}
    \item \textbf{Entity detection} - the process of identifying and extracting entities from text.
    \item \textbf{Coreference resolution } - the process of finding of all expressions (also called mentions) in the text that refer to the same entity.
    \item \textbf{Relation extraction} - the process of identifying semantic relationships between entities.
\end{enumerate}

Previously, researchers addressed each stage individually, but with the increasing power of LLMs, there's been a shift towards end-to-end systems. \citet{kumar2020} suggest employing two LLM components: one for named entity recognition and another one for both relation classification and direction. Similarly, Grapher \citep{melnyk2021grapher} utilizes a pre-trained LLM for entity extraction and relation prediction. However, these methods require users to provide possible relations. More recent methods like PiVE \citep{han2023pive} and AutoKG \citep{zhu2023llms} use LLM prompting strategies for KG construction without additional user input.

The aforementioned methods do not make use of some of the emergent abilities of LLMs, such as in-context learning and the chain-of thought prompting strategy. We decide to leverage these emergent abilities, and take a simple prompt engineering approach to our KG construction step. The techniques used can be summarised as the following:
\begin{itemize}
    \item \textit{Chain-of-thought (CoT) prompting strategy.} Providing intermediate reasoning steps in the prompt to enable LLMs to solve more complex tasks.
    \item \textit{In-context learning.} A method of prompt engineering where one provides several task demonstrations within the prompt, circumventing the need for fine-tuning.
\end{itemize}
The final prompt used in our experiments can be found in the Appendix. We highlight to the reader that our KG construction method is not the main contribution of our work, which is rather the \textit{application of KG construction to the hallucination detection problem}. The major benefit of our KG construction approach is its ease of implementation with any LLM. Furthermore, it is less computationally intensive than methods like PiVE, which performs multiple iterations of improvements to the generated KG. Of course, users may conduct the KG construction stage of GraphEval using their method of choice; the experiments in this paper exhibit the capability of a simple prompting strategy.

\section{GraphCorrect: Correction of hallucinations with GraphEval}
\label{hallucinations_correction}

While the primary focus of this work lies in hallucination detection, GraphEval's breakdown of LLM outputs into triples easily allows for its extension to correct hallucinations within the given context. To achieve this, we first identify all triples within the KG that are likely to contain hallucinations (i.e. those with a probability greater than 0.5, if any). We then employ the following two-step procedure on each identified triple:

\begin{itemize}
    \item[] \textbf{Step 1} - Input the given triple along with the context into an LLM to correct for the potential hallucinations within the triple. This results in a newly generated corrected triple.
    \item[] \textbf{Step 2} - Input the identified triple, its corrected counterpart and the initial LLM output. Selectively replace the information from the original (hallucination-containing) triple with the information from the new triple in the initial LLM output.
\end{itemize}

We name this LLM hallucination correction method as GraphCorrect. The final prompts used in our experiments for both step 1 and step 2 can be found in the Appendix \ref{hallucination_correction_1} and \ref{hallucination_correction_2} respectively. This systematic approach to hallucination correction offers several benefits. First, it tackles each identified hallucination separately, increasing the chances of all perceived hallucinations being corrected. Furthermore, it offers the advantage of exclusively altering the segments of the original text that are suspected to contain a hallucination, leaving other elements untouched and ensuring overall high similarity with the original text. 
Finally, breaking down the entire process into intermediate steps ensures that the original context and the initial LLM output never undergo simultaneous processing within an LLM.  This guarantees safeguards against both the addition of extra information and the loss of information in the LLM output.





\section{Experiments}

\subsection{Benchmarks}
\label{benchmarks}

We conducted two sets of experiments: one focusing on hallucination detection to highlight GraphEval's performance and another on hallucination correction to showcase the advantages of GraphCorrect. For both scenarios, we utilized the SummEval \citep{Fabbri2020SummEvalRS}, QAGS-C and QAGS-X \citep{wang-etal-2020-asking} benchmarks - currently the most prevalent benchmarks in relevant academic literature. All three are concerned with detecting hallucinations in LLM-generated summaries and are human-annotated for factual consistency with respect to the grounding context. Table \ref{tab:benchmarks} contains some statistics pertaining to each of these datasets.


\paragraph{SummEval} The SummEval dataset consists of human evaluations on 16 summarization model outputs from 100 articles from the CNN/DailyMail dataset \citep{hermann2015teaching}. Each summary is labelled on a Likert scale from 1-5 on 4 categories: consistency, coherence, fluency and relevance. We follow the TRUE benchmark \citep{honovich-etal-2022-true} in taking the consistency scores and mapping a score of 5 to being fully consistent, and anything lower to being inconsistent.

\paragraph{QAGS} The QAGS-C and QAGS-X datasets are built from the CNN/DailyMail and the XSum \citep{narayan-etal-2018-dont} datasets, respectively. The human annotators examined the summaries one sentence at a time, and determined the factual consistency of each sentence comparing it to the original article. Three annotators assessed each sentence and the majority decision was recorded. Again, we follow the TRUE benchmark in considering a summary to be factually consistent if and only if all sentences are considered consistent.



\subsection{NLI models in GraphEval}

As mentioned in Section \ref{our_method}, we employ NLI models to perform the second stage of GraphEval - checking the consistency of each individual triple with respect to the context. We conduct experiments using the three most popular NLI-based hallucination detection models available on HuggingFace \footnote{\url{https://huggingface.co}}.

\paragraph{HHEM} Based on the DeBERTaV3 model \citep{he2021deberta} and initially trained on NLI data, the hallucination evaluation model created by Vectara \footnote{\url{https://huggingface.co/vectara/hallucination_evaluation_model}} is further fine-tuned on datasets annotated for consistency. The datasets used for fine tuning were: FEVER \citep{Thorne19FEVER2}, Vitamin C \citep{schuster-etal-2021-get} and PAWS \citep{paws2019naacl}. This model is considerably smaller than the following two models, requiring only 738 MB of memory, and thus has a significantly shorter run-time.

\paragraph{TRUE} The TRUE model is based on a T5-XXL model \citep{2020t5} and is trained similarly to the model described in the TRUE paper \citep{honovich-etal-2022-true}. Instead of the ANLI dataset used in that paper, this model is trained on the same datasets as HHEM, plus the following: SNLI \citep{bowman-etal-2015-large}, MNLI \citep{williams-etal-2018-broad} and Scitail \citep{scitail}. This model requires 45.5 GB of memory.

\paragraph{TrueTeacher} \citet{gekhman2023trueteacher} leverage the ability of LLMs to evaluate hallucinations by generating synthetic data through annotating model-generated summaries. They then use this synthetic data to further fine-tune the model from \cite{honovich-etal-2022-true}, leading to state-of-the-art performance on the TRUE benchmark. This model is the same size as the TRUE model.




\subsection{Experimental settings}
In all experiments conducted in this study necessitating the utilization of an LLM, we use Claude 2 \footnote{\url{https://www.anthropic.com/news/claude-2}}, an LLM from Anthropic, through the Amazon Bedrock API \footnote{\url{https://aws.amazon.com/bedrock/claude/}}. We use the default settings for the LLM: \texttt{temperature} = 1, \texttt{top\_p} = 1, \texttt{top\_k} = 250. We also refer the reader to the Appendix for the prompts used in this work.


\subsection{Results}
\label{results}


\subsubsection{Hallucination detection with GraphEval}

We present our results of hallucination detection for the three NLI models, and their GraphEval counterparts, in Table \ref{tab:auc-roc}. We report the balanced accuracy as our evaluation metric, which corrects for the class imbalance in the SummEval benchmark.
In the case of using the NLI model directly, we classify the example as containing a hallucination if the NLI model returns a probability of more than 0.5. When combining the NLI model with GraphEval, we classify the example as containing a hallucination if at least one triple fed to the NLI model returns a probability of more than 0.5. We see that adding the GraphEval pre-processing step to each of the NLI models almost always improves the balanced accuracy score, sometimes by a considerable amount, such as the results for the SummEval and QAGS-C benchmarks in Table \ref{tab:auc-roc}. On average (weighting by the number of samples in each dataset), adding the GraphEval pre-processing step improves the balanced accuracy by 6.2 (SE=1.3).

\bgroup
\begin{table}[t]
    \centering
    \resizebox{\linewidth}{!}{%
    \begin{tabular}{lccc}
    \hline
         & \textbf{SummEval} & \textbf{QAGS-C} & \textbf{QAGS-X} \\
    \hline
    HHEM & 66.0 & 63.5 & 75.5 \\
    HHEM + GraphEval  & 71.5 & 72.2 & 75.2 \\
    \hline
    TRUE & 61.3 & 61.8 & 72.6 \\
    TRUE + GraphEval & 72.4 & 71.7 & 73.9 \\
    \hline
    TrueTeacher & 74.9 & 75.6 & 79.0 \\
    TrueTeacher + GraphEval & \textbf{79.2} & \textbf{78.1 }& \textbf{79.6} \\
    \end{tabular}
    }
    \vspace{2pt}
    \caption{Balanced accuracy scores for hallucination detection of NLI models (HHEM, TRUE, TrueTeacher) and their GraphEval counterparts on the SummEval, QAGS-C and QAGS-X benchmarks.}
    \label{tab:auc-roc}
\end{table}
\egroup

\bgroup
\def\arraystretch{1.3}
\begin{table*}[b]
    \small
    \centering
    \resizebox{\textwidth}{!}{%
    \begin{tabular}{llcccccc}
    \hline
    \multirow{2}{*}{\textbf{Detection}} & \multirow{2}{*}{\textbf{Dataset}} & \multicolumn{2}{c}{\textbf{ROUGE-1}} & \multicolumn{2}{c}{\textbf{ROUGE-2}} & \multicolumn{2}{c}{\textbf{ROUGE-L}} \\
    & & \textbf{Direct Prompt} & \textbf{GraphCorrect} & \textbf{Direct Prompt} & \textbf{GraphCorrect} & \textbf{Direct Prompt} & \textbf{GraphCorrect} \\
    \hline
    \multirow{3}{*}{HHEM + GraphEval} & SummEval & 0.827 & \textbf{0.915} & 0.772 & \textbf{0.879} & 0.796 & \textbf{0.910} \\
     & QAGS-C & 0.800 & \textbf{0.893} & 0.735 & \textbf{0.841} & 0.769 & \textbf{0.885} \\
     & QAGS-X & 0.649 & \textbf{0.821} & 0.495 & \textbf{0.734} & 0.606 & \textbf{0.815} \\
    \hline
    \multirow{3}{*}{TRUE + GraphEval} & SummEval & 0.781 & \textbf{0.880} & 0.707 & \textbf{0.833} & 0.746 & \textbf{0.871} \\
    & QAGS-C & 0.840 & \textbf{0.894} & 0.780 & \textbf{0.848} & 0.808 & \textbf{0.886} \\
    & QAGS-X & 0.651 & \textbf{0.805} & 0.505 & \textbf{0.706} & 0.613 & \textbf{0.795} \\
    \hline
    \multirow{3}{*}{TrueTeacher + GraphEval} & SummEval & 0.781 & \textbf{0.884} & 0.703 & \textbf{0.839} & 0.737 & \textbf{0.876} \\
     & QAGS-C & 0.809 & \textbf{0.889} & 0.743 & \textbf{0.837} & 0.781 & \textbf{0.881} \\
     & QAGS-X & 0.643 & \textbf{0.797} & 0.486 & \textbf{0.694} & 0.598 & \textbf{0.784} \\
    \end{tabular}
    }
    \vspace{3pt}
    \caption{Average ROUGE-1, ROUGE-2 and ROUGE-L scores measuring similarity between original and corrected summaries using Direct Prompt and GraphCorrect across different datasets and hallucination detection frameworks.}
    \label{tab:rouge_scores}
\end{table*}
\egroup

We hypothesise that the negligible difference between the base NLI model and the model supplemented with GraphEval for the QAGS-X dataset is due to the average length of the generated text (only 18 words, compared with 49 and 63 for QAGS-C and SummEval respectively, see \ref{tab:benchmarks}). This highlights an important aspect of where the most value can be found in our method. When the LLM output is very short, there are less likely to be multiple facts that need to be checked for consistency (which can easily be done without the use of a KG) and the intricacies of the short sentence might even be lost in the KG construction phase. On the other hand, when the LLM output is very long, current methods struggle to test each individual fact against the context, and this is when GraphEval thrives.

It should be noted that even when the results for GraphEval are comparable to the baseline methods, the benefit of using GraphEval is the identification of the specific triple(s) that are inconsistent with the provided context.

\subsubsection{Hallucination correction with GraphCorrect}

Identifying the particular triple(s) likely to harbor a hallucination enables straightforward correction using GraphCorrect, as described in Section \ref{hallucinations_correction}. For each of the evaluation frameworks proposed here (HHEM + GraphEval, TRUE + GraphEval, and TrueTeacher + GrapEval), we compared GraphCorrect to a basic prompting strategy for hallucination correction, serving as a baseline. The prompt used in this baseline approach, referred to as the Direct Prompt henceforth, is provided in Appendix \ref{hallucination_correction_claude}.

For each framework, we initially identify hallucinations, correct only the LLM outputs suspected of containing hallucinations using either GraphCorrect or Direct Prompt, and then reapply the evaluation framework to detect hallucinations in the corrected LLM outputs. Note that this procedure only allows us to measure what we presume to be corrected hallucinations, given the potential for errors in the evaluation frameworks utilized here. We report the percentage of believed corrected hallucinations in Table \ref{tab:hallucination_correction}. A score of 0\% suggests no corrected hallucinations according to the given framework, while a score of 100\% indicates correction of all hallucinations as per the given framework. GraphCorrect outperforms the prompting strategy proposed here by significantly correcting for more hallucinations on all tasks apart from two related to the QAGS-X dataset. As on the hallucination detection task, we hypothesise these results are correlated with the average length of the text, with GraphCorrect bringing most value in longer texts with a more complex structure to unravel and correct.

\bgroup
\begin{table}[b]
    \centering
    \resizebox{\linewidth}{!}{%
    \begin{tabular}{llcc}
    \hline
    \multirow{2}{*}{\textbf{\footnotesize{Detection \& Evaluation}}} & \multirow{2}{*}{\textbf{\footnotesize{Dataset}}} & \multicolumn{2}{c}{\textbf{\footnotesize{Method for Correction}}} \\
    & & \textbf{\footnotesize{Direct Prompt}} & \textbf{\footnotesize{GraphCorrect}} \\
    \hline
    \multirow{3}{*}{\small{HHEM + GraphEval}} & SummEval & 48.6 & \textbf{55.1} \\
     & QAGS-C & 38.5 & \textbf{58.7} \\
     & QAGS-X & 63.2 & \textbf{69.5} \\
    \hline
    \multirow{3}{*}{\small{TRUE + GraphEval}} & SummEval & 49.6 & \textbf{59.5} \\
    & QAGS-C & 42.7 & \textbf{53.7} \\
    & QAGS-X & \textbf{70.8} & 66.7 \\
    \hline
    \multirow{3}{*}{\small{TrueTeacher + GraphEval}} & SummEval & 53.1 & \textbf{59.8} \\
     & QAGS-C & 47.1 & \textbf{59.6} \\
     & QAGS-X & \textbf{71.1} & 69.3 \\
    \end{tabular}
    }
    \vspace{3pt}
    \caption{Percentage of believed corrected hallucinations using a direct prompting strategy and GraphCorrect on the SummEval, QAGS-C and QAGS-X benchmarks. The hallucinations were first detected by HHEM + GraphEval, TRUE + GraphEval and TrueTeacher + GraphEval respectively, and then corrections were evaluated by the same metric.}
    \label{tab:hallucination_correction}
\end{table}
\egroup

Additionally, as previously stated, GraphCorrect offers the advantage of only modifying the segments of text in the LLM outputs susceptible to hallucinations, while leaving other sections unaltered, thereby maintaining high overall similarity with the original text. This characteristic is illustrated in Table \ref{tab:rouge_scores} by assessing the ROUGE-1, ROUGE-2, and ROUGE-L metrics between the original summaries and the corrected versions for both GraphCorrect and Direct Prompt across all experimental scenarios examined in this study. GraphCorrect systematically generates texts that are closer in similarity to the original LLM outputs compared to its counterpart.

\section{Discussion}
\label{future_work}

Our work focuses on detection of hallucinations in closed-domain tasks, where we are interested only in consistency with respect to the provided context. The GraphEval framework could be extended to open-domain hallucination detection by employing agents, as in AutoKG \citep{zhu2023llms}, to first retrieve relevant external sources as the grounding information to check against. 

We expect that in the near future, more research will be conducted on the construction of KGs from unstructured text, which will provide improvements to the first stage of our procedure and ultimately the evaluation performance. Even as LLMs alone become more powerful, this will continue to contribute to improvements in GraphEval's performance.

We observe that, in the knowledge graph construction phase of our procedure, it is possible that some information loss may occur. However, as shown by the results in Section \ref{results}, our method rarely leads to a reduction in balanced accuracy. Furthermore, when it is comparable to the baseline methods, we have the added explainability of identifying the specific triples where the hallucination has occurred.

We believe our hallucination correction framework (GraphCorrect) shows promise and an interesting avenue for future work. However, the effectiveness of the approach described in this work should be assessed manually, rather than relying on the convoluted use of hallucination evaluation frameworks (which only yield measurements of \textit{believed} corrected hallucinations). 

\section{Conclusion}

We introduce GraphEval, a simple and effective pre-processing step for improving the explainability and performance of LLM hallucination detection metrics. Our method leverages LLM's ability to extract information from unstructured text and construct knowledge graphs, whose triples can be fed into out-of-the-box hallucination detection methods.

We demonstrate that GraphEval in conjunction with state-of-the-art NLI models leads to an average improvement in balanced accuracy of 6.2 (SE=1.3) on three popular hallucination benchmarks. Furthermore, our method indicates which triples, in the KG representation of the LLM output, are inconsistent.
To the best of our knowledge, this is the first application of KGs to an LLM-based hallucination evaluation framework and we believe the success of GraphEval will only grow as KG construction methods also improve.

Finally, we examined the issue of hallucination correction and showed that GraphCorrect can effectively address the majority of hallucinations found in LLM outputs while maintaining extremely high similarity with the original texts.




\bibliography{mybibfile}

\appendix

\section{KG Construction Prompt}
\label{}

\begin{lstlisting}[breaklines=true, basicstyle=\small]
 ("system",
    """
    You are an expert at extracting information in 
    structured formats to build a knowledge graph.
    Step 1 - Entity detection: Identify all entities in the raw text. Make sure not to miss any out. Entities should be basic and simple, they are akin to Wikipedia nodes.
    Step 2 - Coreference resolution: Find all expressions in the text that refer to the same entity. Make sure entities are not duplicated. In particular do not include entities that are more specific versions themselves, e.g. "a detailed view of jupiter's atmosphere" and "jupiter's atmosphere", only include the most specific version of the entity.
    Step 3 - Relation extraction: Identify semantic relationships between the entities you have identified.

    Format: Return the knowledge graph as a list of triples, i.e. ["entity 1", "relation 1-2", "entity 2"], in Python code.
    """,
),
("human",
        "Use the given format to extract information from the following input: <input>{input}</input>.  Skip the preamble and output the result as a list within <python></python> tags.",
),
("human",
        """Important Tips: 
            1. Make sure all information is included in the knowledge graph.
            2. Each triple must only contain three strings! None of the strings should be empty.
            3. Do not split up related information into separate triples because this could change the meaning.
            4. Make sure all brackets and quotation marks are matched.
            5. Before adding a triple to the knowledge graph, check the concatenated triple makes sense as a sentence. If not, discard it.
        """,
),
("human",
        """Here are some example input and output pairs.
        
        ## Example 1.
        Input: 
        "The Walt Disney Company, commonly known as Disney, is an American multinational mass media and entertainment conglomerate that is headquartered at the Walt Disney Studios complex in Burbank, California."
        Output: 
        <python> 
        [["The Walt Disney Company", "headquartered at","Walt Disney Studios complex in Burbank, California"], 
        ["The Walt Disney Company", "commonly known as", "Disney"], 
        ["The Walt Disney Company", "instance of", "American multinational mass media and entertainment conglomerate"]] 
        </python>
        
        ## Example 2.
        Input: 
        "Amanda Jackson was born in Springfield, Ohio, USA on June 1, 1985. She was a basketball player for the U.S. women's team."
        Output: 
        <python> 
        [ ["Amanda Jackson", "born in", "Springfield, Ohio, USA"],
        ["Amanda Jackson", "born on", "June 1, 1985"], 
        ["Amanda Jackson", "occupation", "basketball player"], 
        ["Amanda Jackson", "played for", "U.S. women's basketball team"]] </python>

        ## Example 3.
        Input: 
        "Music executive Darius Van Arman was born in Pennsylvania. He attended Gonzaga College High School and is a human being."
        Output: 
        <python> 
        [ ["Darius Van Arman", "occupation", "Music executive"], 
        ["Darius Van Arman", "born in", "Pennsylvania"], 
        ["Darius Van Arman", "attended", "Gonzaga College High School"], ["Darius Van Arman", "instance of", "human being"]] 
        </python>

        ## Example 4.
        Input: "Italy had 3.6x times more cases of coronavirus than China."
        Output: 
        <python> 
        [ ["Italy", "had 3.6x times more cases of coronavirus than", "China"]]
        </python>
        """,
        ),
\end{lstlisting}

\section{Hallucination correction (step 1)}
\label{hallucination_correction_1}

\begin{lstlisting}[breaklines=true, basicstyle=\small]
    """
    You are an expert at extracting information in structured formats from text.
    The following triple contains factually incorrect information.
    Correct it based on the provided context,
    Important Tips: 
        1. A triple is defined as ["entity 1", "relation 1-2", "entity 2"].
        2. A triple must only contain three strings! None of the strings should be empty.
        3. The concatenated triple must make sense as a sentence.
        4. Only return the corrected triple, nothing else.
        
    <triple>{triple}</triple>
    <context>{context}</context>
    
    Remember, it is important that you only return the corrected triple.
    """
\end{lstlisting}

\section{Hallucination correction (step 2)}
\label{hallucination_correction_2}

\begin{lstlisting}[breaklines=true, basicstyle=\small]
    """
    In the following context, replace the information of the old triple with the information of the new one.
    Do not make any other modification to the context.
    Only return the new context.
    <context>{summary}</context>
    <old_triple>{old_triple}</old_triple>
    <new_triple>{new_triple}</new_triple>
    """
\end{lstlisting}

\section{Hallucination correction without a KG}
\label{hallucination_correction_claude}

\begin{lstlisting}[breaklines=true, basicstyle=\small]
    """
    The following summary contains factually incorrect information. 
    Correct it based on the context, but don't change other parts of the summary. 
    Only return the corrected summary, nothing else.
    <summary>{summary}</summary> 
    <context>{context}</context>
    Remember, do minimal changes to the original summary, don't make it longer and keep as much of it as you can exactly the same.
    """
\end{lstlisting}

\end{document}